\documentclass[conference,a4paper]{APSIPA2021}
\usepackage{multirow}
\usepackage{graphicx}
\usepackage{amsmath}
\usepackage[psamsfonts]{amssymb}
\usepackage{amsxtra}
\usepackage{threeparttable}
\usepackage{subcaption}
\usepackage{cite}
\usepackage{color}

\begin{document}

\title{On the Adversarial Transferability \\ of ConvMixer Models}

\author{%
\authorblockN{%
Ryota Iijima\authorrefmark{1}, 
Miki Tanaka\authorrefmark{2}, 
Isao Echizen\authorrefmark{3}, and
Hitoshi Kiya\authorrefmark{4}
}
\authorblockA{%
\authorrefmark{1}
Tokyo Metropolitan University, Tokyo, Japan \\
E-mail: iijima-ryota@ed.tmu.ac.jp}
\authorblockA{%
\authorrefmark{2}
Tokyo Metropolitan University, Tokyo, Japan \\
E-mail: tanaka-miki@ed.tmu.ac.jp}
\authorblockA{%
\authorrefmark{3}
National Institute of Informatics (NII), Tokyo, Japan\\
E-mail: iechizen@mi.ac.jp}
\authorblockA{%
\authorrefmark{4}
Tokyo Metropolitan University, Tokyo, Japan \\
E-mail: kiya@tmu.ac.jp}
}

\maketitle
\thispagestyle{empty}

\begin{abstract}
Deep neural networks (DNNs) are well known to be vulnerable to adversarial examples (AEs).
In addition, AEs have adversarial transferability, which means AEs generated for a source model can fool another black-box model (target model) with a non-trivial probability.
In this paper, we investigate the property of adversarial transferability between models including ConvMixer, which is an isotropic network, for the first time. To objectively verify the property of transferability, the robustness of models is evaluated by using a benchmark attack method called AutoAttack.
In an image classification experiment, ConvMixer is confirmed to be weak to adversarial transferability.
\end{abstract}

\section{Introduction}
Deep neural networks (DNNs) have been deployed in many applications including security-critical ones such as biometric authentication and automated driving, but DNNs are vulnerable to adversarial examples (AEs), which are perturbed by noises to mislead DNNs without affecting human perception.
In addition, AEs generated for a source model fool other (target) models, and this property is called adversarial transferability.
This transferability allows attackers to use a substitute model to generate AEs that may also fool other target models, so reducing its influence has become an urgent issue.  \par

Many studies have investigated both AEs and the transferability of AEs to build models robust against these attacks \cite{aprilpyone2021block, croce2020reliable, croce2020minimally, maksym2020square}.
In contrast, various methods for generating AEs have also been proposed to fool DNNs \cite{kiya2022overview, ian2015explaining, su2019one}.
In addition, recently, the adversarial transferability between the vision transformer (ViT) \cite{Alexey2021an} and convolutional neural network models was confirmed to be low \cite{mahmood2021on, naseer2022on, tanaka2022transferability}.
However, the adversarial transferability of ConvMixer \cite{trockman2022patches} has never been investigated.
ConvMixer is an isotropic network the same as ViT, but it includes CNNs. Accordingly, in this paper, we aim to investigate the adversarial transferability between ConvMixer and other CNN models to confirm the difference between ConvMixer and ViT models. \par

In this paper, the adversarial transferability of encrypted models is evaluated under the use of a benchmark attack method, referred to as AutoAttack, which was proposed to objectively evaluate the robustness of models against AEs.
In an experiment, the use of ConvMixer models is verified not only to make the transferability of AEs generated from other CNN models weak but to also make that generated from ViT models weak.

\section{Related Work}

\subsection{Adversarial Examples}
AEs are classified into three groups based on the knowledge of a particular model and training data available to the adversary: white-box, black-box, and gray-box.
Under white-box settings \cite{ian2015explaining, madry2018towards, croce2020minimally}, the adversary has direct access to the model, its parameters, training data, and defense mechanism.
However, the adversary does not have any knowledge on the model, except the output of the model in black-box attacks \cite{su2019one, maksym2020square, yandogn2019nattack}.
Situated between white-box and black-box methods are gray-box attacks that imply that the adversary knows something about the system.
With the development of AEs, numerous adversarial defenses have been proposed in the literature.
Conventional defenses have been compared under a benchmark attack framework called AutoAttack \cite{croce2020reliable}. \par

However, conventional defenses are not effective against adversarial transferability, which means AEs generated for a source model can fool another black-box model (target model) with a non-trivial probability in general, although many studies \cite{nicolas2016transfer, christian2014intriguing, liu2017delving, mahmood2021on} have investigated adversarial transferability.
As mentioned above, the adversarial transferability between the vision transformer (ViT) and convolutional neural network models was recently confirmed to be low.
This result is expected to lead to a novel insights into defending models.
However, the adversarial transferability of ConvMixer has never been investigated.
ConvMixer is an isotropic network the same as ViT, but it includes CNNs.
Accordingly, in this paper, we aim to investigate the adversarial transferability between ConvMixer and other CNN models to confirm the difference between ConvMixer and ViT models.

\subsection{AutoAttack}
Many defenses against AEs have been proposed, but it is very difficult to judge the value of defense methods without an independent test. For this reason, AutoAttack \cite{croce2020reliable}, which is an ensemble of adversarial attacks used to test adversarial robustness objectively, was proposed as a benchmark attack.
AutoAttack consists of four attack methods: Auto-PGD-cross entropy (APGD-ce) \cite{croce2020reliable}, APGD-target (APGD-t), FAB-target (FAB-t) \cite{croce2020minimally}, and Square Attack \cite{maksym2020square}, as summarized in Table I.
In this paper, we use these four attack methods to objectively evaluate the transferability of AEs.

\begin{table}[h]
    \centering
    \caption{Attack methods used in AutoAttack}
    \begin{tabular}{c|c|c|}
    \multirow{2}{*}{Attack} & Target (T) & White-box (W) \\
    & / Non-target (N) & / Black-box (B) \\
    \hline
    APGD-ce & N & W \\
    APGD-t & T & W \\
    FAB-t & T & W \\
    Square & N & B \\
    \end{tabular}
    \label{table1}
\end{table}

\subsection{ConvMixer}
ConvMixer \cite{trockman2022patches} is well-known to perform highly in image classification tasks, even though it has a small number of model parameters.
It is a type of isotropic network.
It is inspired by the vision transformer (ViT) \cite{Alexey2021an}, so the architecture has a unique feature, called patch embedding. \par

Fig. 1 shows the architecture of the network, which consists of two main structures: patch embedding and ConvMixer layers.
First, an input image $x$ is divided into patches by patch embedding with a patch size of $P$.
Next, the patches are transformed by $L$ ConvMixer layers.
Each layer consists of depthwise convolution and pointwise convolution.
Therefore, ConvMixer is a CNN-based model, not a Transformer-based model like ViT.
Finally, the output of the $L$th ConvMixer layer is transformed by Global Average Pooling and a softmax function to obtain a result. \par

In previous work, the transferability between CNN models and ViT was mentioned to be lower than the transferability between CNN models \cite{mahmood2021on}.
However, the reason for the low transferability was not clearly indicated.
In this paper, we evaluate the hypothesis that patch embedding is the cause of the low transferability.
To evaluate this hypothesis, we use ConvMixer as a CNN-based model with patch embedding, and we compare this model with a transformer-based model, ViT.
\begin{figure*}[t]
    \centering
    \includegraphics[keepaspectratio, width=160mm]{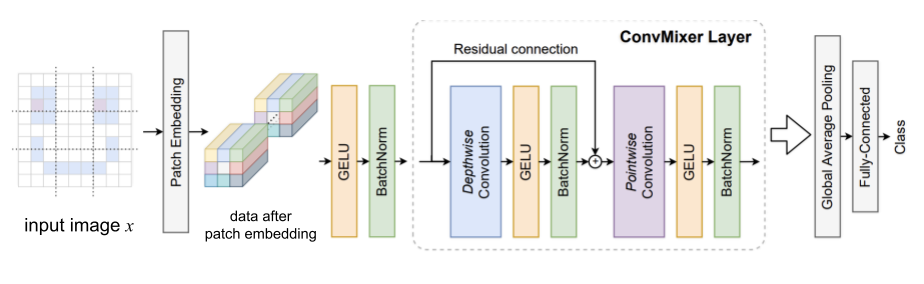}
    \caption{Architecture of ConvMixer \cite{trockman2022patches}}
    \label{fig:my_label}
\end{figure*}

\section{Evaluation of Adversarial Transferability}
In this paper, robustness against adversarial transferability is evaluated under the use of ConvMixer models.
Targets for comparison are summarized here.

\subsection{Type of Model}
Various models have been proposed for image classification tasks. The residual network (ResNet) \cite{he2016deep} and very deep convolutional network (VGGNet) \cite{karen2015deep} use a convolutional neural network(CNN).
In contrast, vision transformers (ViT) \cite{Alexey2021an} do not use CNNs.
As mentioned, the transferability between CNN models and ViT was found to be lower than the transferability between CNN models \cite{mahmood2021on}.
In this paper, we use three CNN models, ResNet18, ResNet50 and VGG16, and we also use two isotropic networks, ViT and ConvMixer to investigate the transferability of AEs between models. In addition, encrypted ConvMixer models, which were proposed for defense against AEs \cite{aprilpyone2021block}, are evaluated in terms of the transferability of AEs. 

\subsection{Encrypted ConvMixer Model}
A block-wise transformation with secret keys, which was inspired by learnable encryption \cite{maung2021ensemble, kiya2022overview, tanaka2018learnable,Madono2020BlockwiseSI, chuman2019encryption,watanabe2004afast,ibuki2016unitary, sirichotedumrong2021a}, was proposed for adversarial defense \cite{aprilpyone2021block} where a model is trained by using encrypted images as below (see Fig. 2).
\begin{enumerate}
    \item Each training image x is divided into blocks with a size of $M \times M$.
    \item Every block in x is encrypted by a transformation algorithm with secret keys to generate encrypted images.
    \item A model is trained by using the encrypted images to generate an encrypted model.
    \item A query image is encrypted with key K, and the encrypted image is then input to the encrypted model to get an estimation result.
\end{enumerate}
There are two parameters used in steps 1) and 2) when encrypting a model: the block size $M$ and transformation algorithm. In \cite{aprilpyone2021block}, three transformation algorithms were proposed: pixel shuffling (SHF), bit flipping (NP), and format-preserving, Feistel-based encryption (FFX). \par

Fig. 2 shows an example of images encrypted from the original image with a size of $224 \times 224$ in Fig. 2 (a) by using these three algorithms with $M=16$.
In this paper, we evaluate the transferability of AEs between models including encrypted ones.

\begin{figure}[h]
    \centering
    \begin{minipage}{0.45\linewidth}
    \centering
    \includegraphics[keepaspectratio, scale=0.4]{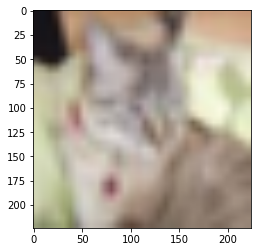}
    \subcaption{}
    \end{minipage}
    \begin{minipage}{0.45\linewidth}
    \centering
    \includegraphics[keepaspectratio, scale=0.4]{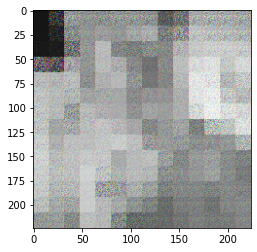}
    \subcaption{}
    \end{minipage} \\
    \begin{minipage}{0.45\linewidth}
    \centering
    \includegraphics[keepaspectratio, scale=0.4]{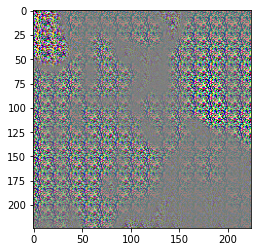}
    \subcaption{}
    \end{minipage}
    \begin{minipage}{0.45\linewidth}
    \centering
    \includegraphics[keepaspectratio, scale=0.4]{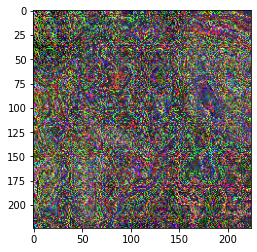}
    \subcaption{}
    \end{minipage}
    \caption{Example of transformed images ($M=16$) (a): plain image, (b): SHF, (c): NP, (d): FFX}
    \label{fig4}
\end{figure}

\subsection{AEs Designed with Source Model}
Fig. 3 shows the framework of attacks with adversarial transferability where AEs are designed by using a source model.
In this paper, the four methods used in AutoAttack are used to generate AEs. After generating AEs, the AEs are input to a target model. \par

In this paper, we consider not only plain models but also encrypted models as a source model.
Adversarial defenses with encrypted models are illustrated in Fig. 4 \cite{aprilpyone2021block} where an encrypted model is trained with encrypted images, and an AE is transformed with a secret key.
For the case of using encrypted models, the framework of attacks with adversarial transferability is given in Fig. 5 where a perturbation generated from an encrypted model is added to plain query images.




\begin{figure}
    \centering
    \includegraphics[keepaspectratio, scale=0.28]{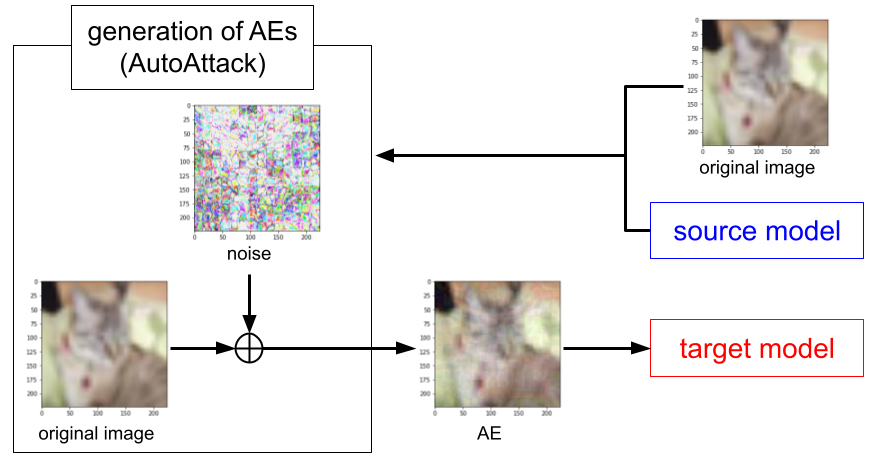}
    \caption{Framework of attacks with adversarial transferability}
    \label{fig:my_label}
\end{figure}

\begin{figure}
    \centering
    \includegraphics[keepaspectratio, scale=0.4]{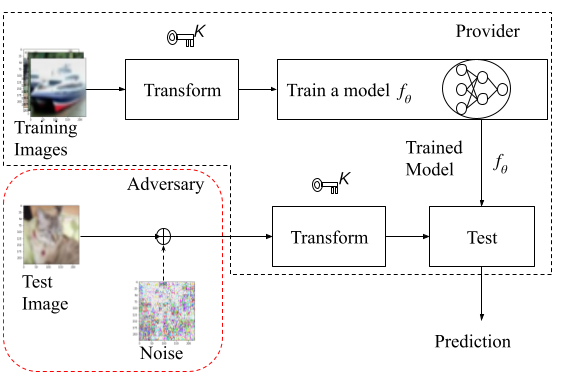}
    \caption{Adversarial defense with encrypted models \cite{aprilpyone2021block}}
    \label{fig:my_label}
\end{figure}

\begin{figure}
    \centering
    \includegraphics[keepaspectratio, scale=0.28]{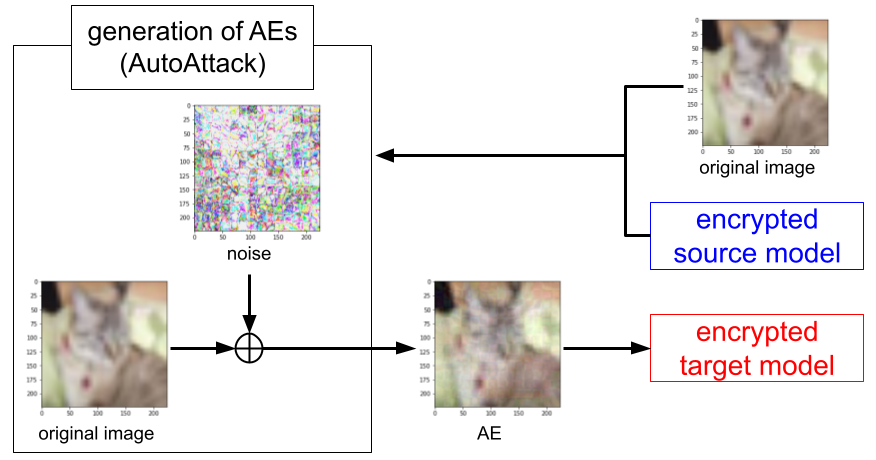}
    \caption{Framework of attacks with adversarial transferability (with encrypted models)}
    \label{fig:my_label}
\end{figure}

\section{Experiment}
\subsection{Experiment Setup}
In the experiment, we used five networks for image classification, ResNet18, ResNet50, VGG16, ViT, and ConvMixer, to evaluate the transferability of AEs.
In addition, ConvMixer was also used for generating encrypted models where the above three transformation algorithms,  SHF, NP, and FFX, were applied to images in accordance with the steps in Sec I\hspace{-.1em}I\hspace{-.1em}I B.
The experiment was carried out on the CIFAR-10 dataset (with 10 classes), which consists of 60,000 color images with a dimension of $3 \times 32 \times 32$, where 50,000 of the images are for training, 10,000 images are for testing, and each class contains 6000 images.
The images in the dataset were resized to $3 \times 224 \times 224$ for fitting with pretrained ViT models.
AEs were generated by using the four attack methods used in AutoAttack under the $l_\infty$ norm with $\epsilon = 8/255$: APGD-ce, APGD-t, FAB-t, and Square. \par

The transferability of the AEs was evaluated by using the attack success rate (ASR).
The ASR between a source classifier model $\mathrm{C_s}$ and a target classifier model $\mathrm{C_t}$ is given by
\begin{equation}
    ASR = \frac{100}{N_\mathrm{c}} \sum^{N_\mathrm{c}}_{k=1} \left \{
    \begin{aligned}
    1 \: &(\mathrm{A_{C_t}}(x_k, y_k) \land \left \{ \mathrm{C_s}(x_k) = y_k \right \} ) \\
    0 \: & (\mathrm{otherwise})
    \end{aligned},
    \right .
\end{equation}
\begin{equation}
    \mathrm{A_{C}}(x, y) = \left \{ \mathrm{C}(x) = y \right \} \land \left \{ \mathrm{C}(x^\mathrm{adv}) \neq y \right \} ,
\end{equation}
where $N_\mathrm{c}$ is the number of images correctly classified in both $\mathrm{C_s}$ and $\mathrm{C_t}$, $x_k$ is an image used to generate an AE, $y_k$ is a label of an image $x_k$, and $x^\mathrm{adv}$ is an AE generated from an image $x$.
The ASR is in the range of $[0, 100]$, and a lower value indicates that the transferability is lower.

\subsection{Results}

\subsubsection{Transferability among Five Models}
Tables I\hspace{-.1em}I and I\hspace{-.1em}I\hspace{-.1em}I show the transferability among the five models in terms of ASR.
In Table I\hspace{-.1em}I, the result when ResNet18 was chosen as the source model is given.
From the table, ResNet50 and VGG16 were misled by AEs.
In contrast, ViT and ConvMixer were not misled.
Table I\hspace{-.1em}I\hspace{-.1em}I shows the ASR when ConvMixer was used as the source model.
The AEs generated for ConvMixer could not mislead the other four models.
Both ViT and ConvMixer are isotropic networks, but ViT was not fooled.
Accordingly, the transferability between ConvMixer and other models including ViT was confirmed to be low.



\begin{table}[t]
    \centering
    \caption{ASR of five models (Source: ResNet18)}
    \begin{tabular}{c||c|c|c|c|}
    Target & APGD-ce & APGD-t & FAB-t & Square \\
    \hline
    ResNet18 & 100.0 & 100.0 & 99.62 & 100.0\\
    ResNet50 & 97.08 & 68.97 & 1.35 & 21.74\\
    VGG16 & 54.39 & 33.06 & 0.69 & 7.68\\
    ViT & 32.16 & 5.49 & 0.04 & 4.09\\
    ConvMixer & 17.19 & 8.33 & 0.39 & 5.69\\
    \end{tabular}
    \label{table1}
\end{table}

\begin{table}[t]
    \centering
    \caption{ASR of five models (Source: ConvMixer)}
    \begin{tabular}{c||c|c|c|c|}
    Target & APGD-ce & APGD-t & FAB-t & Square \\
    \hline
    ResNet18 & 48.91 & 27.65 & 0.5 & 7.0\\
    ResNet50 & 47.64 & 26.3 & 0.5 & 10.41\\
    VGG16 & 40.88 & 27.17 & 0.63 & 5.21\\
    ViT & 36.48 & 10.6 & 0.09 & 3.14\\
    ConvMixer & 100.0 & 100.0 & 99.99 & 100.0\\
    \end{tabular}
    \label{table1}
\end{table}

\subsubsection{Transferability between ConvMixer and encrypted ConvMixer}
The transferability between ConvMixer and encrypted ConvMixer is shown in Table I\hspace{-.1em}V where ConvMixer was used as the source model, and encrypted ConvMixer models were used as target models.
From the result, the use of encrypted models enhanced the robustness against each attack.
In particular, models encrypted by using FFX was effective compared to other encryption methods.
In addition, the transferability was not affected by selection of block sizes.

\begin{table}[t]
    \centering
    \caption{ASR of ConvMixer models (Source: ConvMixer, Target: Encrypted ConvMixer)}
    \begin{tabular}{c|c||c|c|c|c|}
    transform & block size & APGD-ce & APGD-t & FAB-t & Square \\
    \hline
    SHF & 16 & 71.13 & 38.41 & 2.22 & 17.4\\
    SHF & 8 & 72.16 & 38.55 & 1.83 & 17.04\\
    SHF & 4 & 74.95 & 42.51 & 1.85 & 16.45\\
    \hline
    NP & 16 & 73.07 & 39.94 & 1.84 & 17.33\\
    NP & 8 & 73.78 & 39.8 & 1.77 & 16.13\\
    NP & 4 & 72.57 & 39.36 & 1.85 & 17.64\\
    \hline
    FFX & 16 & 30.59 & 18.5 & 2.7 & 16.55\\
    FFX & 8 & 31.15 & 19.31 & 2.87 & 15.21\\
    FFX & 4 & 30.74 & 18.49 & 2.81 & 14.53\\
    \end{tabular}
    \label{table1}
\end{table}

\subsubsection{Transferability between encrypted ConvMixer models}
Table V shows the transferability between models encrypted under different conditions, where the source model and target models were also encrypted.
The source model was encrypted by using SHF with a block size of 16.
As shown in the table, using different encryption parameters had no ability to reduce the transferability.
Conversely, from Table I\hspace{-.1em}V and V, the transferability between encrypted ConvMixer models was higher than that between plain ConvMixer and encrypted ConvMixer ones.

\begin{table*}[t]
    \centering
    \caption{ASR of ConvMixer models (Source: Encrypted ConvMixer (SHF, $M=16$), Target: Encrypted ConvMixer (with key different from key used for source model))}
    \begin{tabular}{c|c||c|c|c|c|}
    transform & block size & APGD-ce & APGD-t & FAB-t & Square \\
    \hline
    SHF (with same & \multirow{2}{*}{16} & \multirow{2}{*}{100.0} & \multirow{2}{*}{100.0} & \multirow{2}{*}{100.0} & \multirow{2}{*}{100.0} \\
    key as source) &  &  & &  & \\
    SHF & 16 & 94.61 & 71.84 & 1.82 & 16.05\\
    SHF & 8 & 95.46 & 71.48 & 1.94 & 15.55\\
    SHF & 4 & 95.85 & 73.69 & 1.94 & 14.95\\
    \hline
    NP & 16 & 95.76 & 73.26 & 2.12 & 15.64\\
    NP & 8 & 95.66 & 72.64 & 1.88 & 14.57\\
    NP & 4 & 95.13 & 71.79 & 2.02 & 16.0\\
    \hline
    FFX & 16 & 45.42 & 28.97 & 3.16 & 15.87\\
    FFX & 8 & 47.16 & 29.94 & 3.08 & 14.22\\
    FFX & 4 & 45.84 & 29.09 & 3.03 & 13.86\\
    \end{tabular}
    \label{table1}
\end{table*}

\section{Conclusion}
In this paper, we investigated the transferability between models including ConvMixer.
To objectively verify the transferability, the four attack methods used in AutoAttack were used to generate AEs from a source model.
In the experiment, the use of ConvMixer was confirmed to reduce the influence of the transferability between models including ViT.
However, the use of encrypted ConvMixer models could not enhance the effect of reducing the influence of adversarial transferability.
We will consider how the influence of adversarial transferability is avoided as the next step.

\section*{Acknowledgment}
This research was partially supported by JST CREST (Grant Number JPMJCR20D3) and ROIS NII Open Collaborative Research 2022-(22S1401).

\bibliographystyle{ieicetr}
\bibliography{main}

\end{document}